\pdfoutput=1

\documentclass[11pt]{article}

\usepackage{ACL2023}

\usepackage{times}
\usepackage{latexsym}

\usepackage[T1]{fontenc}

\usepackage[utf8]{inputenc}

\usepackage{microtype}

\usepackage{inconsolata}

\usepackage{bbm}
\usepackage{multirow}
\usepackage{booktabs}
\usepackage{graphicx}
\usepackage{pifont}
\newcommand{\xmark}{\ding{55}}%

\usepackage{colortbl}
\usepackage{amsmath}
\usepackage{amssymb}
\usepackage{bm}

\usepackage{cleveref}
\crefname{section}{§}{§§}
\Crefname{section}{§}{§§}

\usepackage{makecell}

\definecolor{darkgreen}{RGB}{0,160,0}
\definecolor{orange}{RGB}{238,118,0}

\makeatletter
\newcommand{\ssymbol}[1]{$^{\@fnsymbol{#1}}$}
\makeatother

\title{MultiCapCLIP: Auto-Encoding Prompts for Zero-Shot\\Multilingual Visual Captioning}

\author{Bang Yang\textsuperscript{1,2}\thanks{~~Equal contribution.}, Fenglin Liu\textsuperscript{3}\footnotemark[1], Xian Wu\textsuperscript{4}, Yaowei Wang\textsuperscript{2}, Xu Sun\textsuperscript{5}\thanks{~~Corresponding authors.}, and Yuexian Zou\textsuperscript{1}\footnotemark[2]\\
\textsuperscript{1}ADSPLAB, School of ECE, Peking University \ \ \textsuperscript{2}Peng Cheng Laboratory\\
\textsuperscript{3}University of Oxford \ \ \textsuperscript{4}Tencent Jarvis Lab\ \ \textsuperscript{5}School of Computer Science, Peking University\\
\tt \{yangbang, zouyx\}@pku.edu.cn; fenglin.liu@eng.ox.ac.uk \\
}

\begin{document}
\maketitle
\begin{abstract}
Supervised visual captioning models typically require a large scale of images or videos paired with descriptions in a specific language (i.e., the vision-caption pairs) for training. However, collecting and labeling large-scale datasets is time-consuming and expensive for many scenarios and languages. Therefore, sufficient labeled pairs are usually not available. To deal with the label shortage problem, we present a simple yet effective zero-shot approach MultiCapCLIP that can generate visual captions for different scenarios and languages without any labeled vision-caption pairs of downstream datasets. In the training stage, MultiCapCLIP only requires text data for input. Then it conducts two main steps: 1) retrieving concept prompts that preserve the corresponding domain knowledge of new scenarios; 2) auto-encoding the prompts to learn writing styles to output captions in a desired language. In the testing stage, MultiCapCLIP instead takes visual data as input directly to retrieve the concept prompts to generate the final visual descriptions. The extensive experiments on image and video captioning across four benchmarks and four languages (i.e., English, Chinese, German, and French) confirm the effectiveness of our approach. Compared with state-of-the-art zero-shot and weakly-supervised methods, our method achieves 4.8\% and 21.5\% absolute improvements in terms of BLEU@4 and CIDEr metrics. Our code is available at \url{https://github.com/yangbang18/MultiCapCLIP}.
\end{abstract}

\section{Introduction}

Visual captioning targets to first 1) understand the information of visual inputs, which are typically videos or images, and then 2) produces a corresponding textual sentence describing the visual objects/attributes/relationships. 
Visual captioning has drawn remarkable attention from natural language processing and computer vision fields due to its wide applications, e.g., cross-modal retrieval \cite{luo2022clip4clip,cheng2023ssvmr} and helping the visually impaired \cite{ccayli2021mobile}. 
Currently, visual captioning models based on the encoder-decoder framework \cite{huang2020image,liu2020prophet,yang2021non,zhang2021open,hu2022scaling,lin2022swinbert} have achieved tremendous progress in advancing the state-of-the-art.
These models are usually trained with full supervision and rely on large-scale humanly-annotated training data (i.e., vision-caption pairs), which needs expensive labeling work. 
In particular, when it comes to Non-English caption systems, it is challenging to collect and label sufficient vision-caption pairs in a timely manner, which prevents such encoder-decoder models from rapid deployment in different scenarios and languages.

\begin{figure*}[t]
\centering
\includegraphics[width=0.97\linewidth]{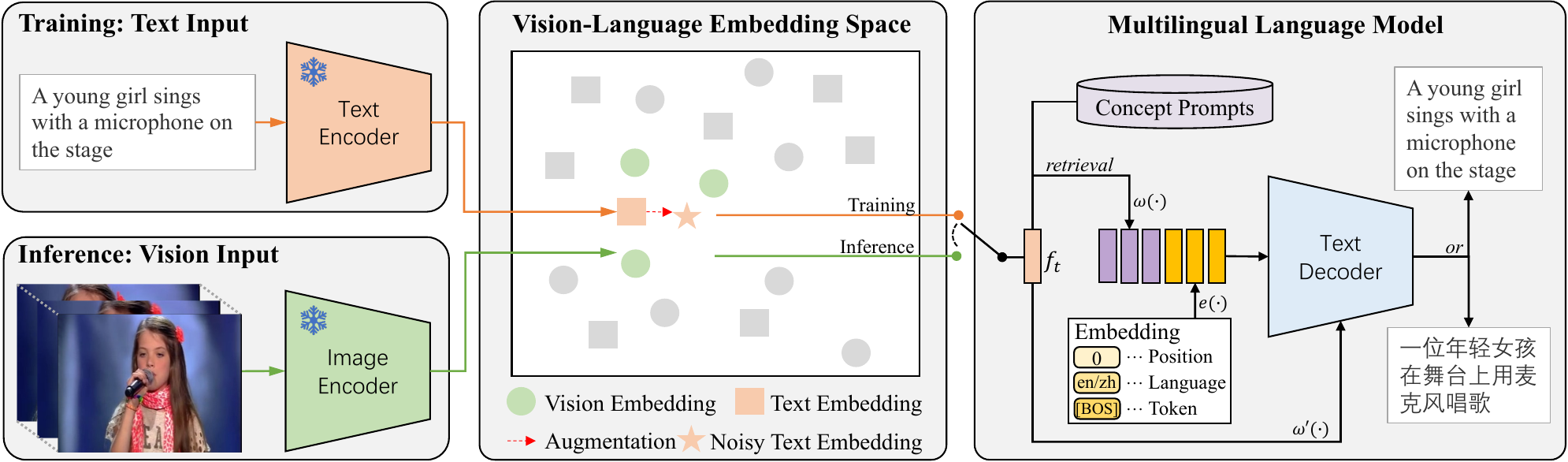}
\caption{Illustration of MultiCapCLIP. It comprises of a frozen CLIP-like vision-language pre-trained model and a trainable multilingual language model. During training, MultiCapCLIP only requires text data and aims to produce a reconstruction/translation result based on the text feature $f_t$ of the source input and its relevant concept prompts (\cref{sec:concept_prompts}). During inference, MultiCapCLIP replaces $f_t$ with the feature(s) of an image or a video for captioning.}
\label{fig:framework}
\end{figure*}

To deal with the shortage of labeled pairs, we propose the MultiCapCLIP - a prompt-based natural language auto-encoder. As shown in Figure~\ref{fig:framework}, MultiCapCLIP only requires textual input for training, and it can conduct zero-shot multilingual visual captioning, including image and video captioning. Therefore, MultiCapCLIP can deal with the situation where the labeled vision-caption pairs are missing. 
MultiCapCLIP is particularly suitable for new scenarios and languages, improving the practical value of visual captioning.

To implement MultiCapCLIP, we first adopt a pre-trained vision-language model, i.e., CLIP \cite{radford2021learning}, as our backbone. CLIP has shown success in correlating the visual and textual modalities into the same latent space (vision-language embedding space) \cite{tewel2022zerocap,su2022language,zeng2023socratic}.
We observe two critical issues for zero-shot visual captioning: the understanding of domain visual knowledge (e.g., objects, attributes, and relationships) and the generation of descriptive sentences in a specific writing style and language. 
Therefore, we propose a prompt-based auto-encoder, which introduces the visual concept prompts $\mathcal{P}$ to preserve the corresponding domain knowledge and writing styles of zero-shot visual captioning.
During training, given the text-only data, we train the model by reconstructing the caption $S$ in the $S \to \mathcal{P} \to S$ auto-encoding pipeline. 
Since the auto-encoding process reconstructs the same input sentence, the model training needs only unlabeled text data. In the reconstruction process, the model is able to preserve the necessary domain knowledge and the writing styles of visual captioning~\cite{Wang2016Auto,Tschannen2018Recent}.
During inference, we can directly take the vision input $V$ as queries to retrieve the domain knowledge preserved in the visual concept prompts and finally rely on the learned writing styles in a specific language in the text decoder to generate visual descriptions in the $V \to \mathcal{P} \to S$ pipeline.

Meanwhile, to further bridge the modality gap between the visual and textual data \cite{liang2022mind}, we introduce an augmentation method, including input augmentation and feature augmentation, which can boost the robustness of the model and in turn improve the performance of zero-shot visual captioning.
The experiments on four benchmark datasets, i.e., MS-COCO \cite{chen2015microsoft}, MSR-VTT \cite{xu2016msr}, VATEX \cite{wang2019vatex}, and Multi30K \cite{elliott-etal-2016-multi30k}, show that our approach can accurately and data-efficiently generate visual captions in English, Chinese, German, and French.


Overall, our main contributions are as follows:

\begin{itemize}
    \item We propose a simple yet effective approach MultiCapCLIP that requires no downstream labeled data to make the first attempt for zero-shot multilingual visual captioning.

    \item MultiCapCLIP first introduces visual concept prompts to preserve the domain knowledge and then auto-encodes them to learn the writing styles of captioning. After text-only training, our approach can shift from text-to-text generation to vision-to-text generation.

    \item The out-of-domain and in-domain experiments on image and video captioning across different languages show that our approach trained on text-only data significantly outperforms previous zero-shot/weakly-supervised methods trained on unpaired or partial labeled visual and textual data, setting new state-of-the-art zero-shot performance.
\end{itemize}

\section{Approach}
In this section, we first give a brief review of CLIP, whose vision-language embedding space lays a foundation for our approach. Next, we introduce the framework of the proposed MultiCapCLIP, followed by two key components: concept prompts and textual augmentations.

\subsection{A Brief Review of CLIP}
\label{sec:review_of_clip}

CLIP uses two independent encoders to process image and text input separately and then bridges the gap between modalities with contrastive learning. The image encoder $\phi_v(\cdot)$ can be a convolutional neural network like ResNet \cite{DBLP:conf/cvpr/HeZRS16} or a vision Transformer like ViT \cite{dosovitskiy2021an}, and it extracts a feature vector for each input image. The text encoder $\phi_t(\cdot)$ is based on Transformer \cite{Vaswani2017transformer}, and it outputs a vector representation of the input text. 
By training two encoders on 400M image-text data with noisy correspondences under the InfoNCE objective \cite{oord2018representation}, CLIP learns a powerful vision-language embedding space that measures image-text similarity well and enables open-vocabulary classification. In this paper, we re-purpose CLIP for zero-shot multilingual visual captioning and always keep $\phi_v(\cdot)$ and $\phi_t(\cdot)$ frozen.

\subsection{Overview of MultiCapCLIP}
As shown in Figure~\ref{fig:framework}, MultiCapCLIP consists of visual and textual encoders from CLIP and a trainable Multilingual Language Model (MLM). MultiCapCLIP supports English text\footnote{The training corpora for CLIP is mainly in English.}, images or videos as inputs and can produce output in desired language. Specifically, we implement MLM with a stack of Transformer decoder blocks, each of which comprises a masked self-attention layer, a cross-attention layer, and a feed-forward layer. Moreover, we add explicit signals in the embedding layer to indicate which language to be generated.

Let denote the text input as $S$, the vision input as $V$, and concept prompts as $P$. Unlike typical visual captioning models that are trained on a vision-text dataset, MultiCapCLIP relies on a text dataset and follows the $S \to P \to S$ auto-encoding pipeline during training. Based on the semantic alignment characteristic of CLIP's feature space, MultiCapCLIP uses the $V \to P \to S$ pipeline for visual captioning during inference. We extend MultiCapCLIP to support multilingual text generation by using parallel corpora with $(S, T)$ pairs, where $T$ denotes the target text in a desired language. In such a case, MultiCapCLIP follows the $S/V \to P \to T$ translation pipeline. 

In the following, we will detail how to extract and leverage $P$ in Section \ref{sec:concept_prompts}. Then in Section \ref{sec:perturbation}, we will introduce an augmentation method to improve the training of MultiCapCLIP.

\subsection{Decoding with Concept Prompts}
\label{sec:concept_prompts}
A set of visual concepts is a good embodiment of domain visual knowledge because a visual concept (e.g., ``a young girl'') manifest as the explicit clue in the vision input. Given a pure text dataset, we use the spaCy toolkit\footnote{\url{https://spacy.io}} to extract noun phrases and reserve the most frequent 1,000 noun phrases as visual concepts, which are first embedded into a prompt template ``\{concept\}''\footnote{The simplest prompt template ``\{concept\}'' produced better performance than other templates like ``a concept of \{concept\}'' in our preliminary experiments.} and then fed into the CLIP's text encoder $\phi_t(\cdot)$ to extract l2-normalized concept features $C=\{c_1, \dots, c_{1000}\}$.

During training, given the text input $S$, we first encode it into a global feature $f_t$:
\begin{equation}
\label{eq:text_feature}
   f_t = {\rm Norm}(\phi_t(S)),
\end{equation}
where ${\rm Norm}(\cdot)$ denotes L2 normalization. Next, we calculate the dot product of $f_t$ and $C$ to measure cosine similarities, based on which we obtain \emph{soft} concept prompts $P$, a subset of $C$ that includes $K$ concept features most semantically similar to $f_t$. Assuming that the dimension of vectors outputted by CLIP is $d$, $P$ is in the shape of $K*d$. To prompt MLM, we prefix embeddings of the target text $S$ with $P$ to obtain the final input embeddings $E$:
\begin{equation}
\label{eq:embedding}
   E = {\rm Concat}(\omega(P), e(S))
\end{equation}
where $\omega(\cdot)$ is implemented as a fully connected layer followed by a layer normalization (LN) \cite{ba2016layer}, and $e(\cdot)$ denotes the summation of position, language, and token embeddings for each $s_i \in S$, followed by LN. The prompt sequence generated by $\omega(P)$ and token sequence generated by $e(S)$ are concatenated together and sent to the text decoder of MLM to regenerate the input sequence $S$. Considering that $f_t$ may contain information supplementary to $P$, we do not discard $f_t$. We first feed the projected feature $f = \omega'(f_t)$, where $\omega'(\cdot)$ has the same structure as $\omega(\cdot)$ but shares no parameters, into the text decoder of MLM. Then we calculate the cross attention between $f$ and $E$. 
We train the model with a cross-entropy loss:
\begin{equation}
\label{eq:loss}
   \mathcal{L} = - \sum_{i=1}^{|S|} \log p_{\theta}(s=s_i|S_{<i}, P, f_t),
\end{equation}
where $p_{\theta}(\cdot)$ is MLM's predicted distribution over a vocabulary and $\theta$ denotes all trainable parameters.

During inference, we process the vision input $V$ in a similar manner, except that we use CLIP's image encoder $\phi_v(\cdot)$ to obtain $V$'s vector representation $f_v$ and obtain relevant concept prompts $P$ based on (averaged) image-concept similarities. Given the previously generated text $S_{<i}$, the prediction of the next token is based on the following probability distribution:
\begin{equation}
\label{eq:distribution}
   p_{\theta}(s|S_{<i}, P, f_v).
\end{equation}

\subsection{Training with Augmentations}
\label{sec:perturbation}

Our MultiCapCLIP's ability of shifting text-to-text generation to vision-to-text generation is built on the assumption that the paired vision-text data is well aligned in the vision-language embedding space of CLIP. However, \citet{liang2022mind} demonstrated that there exists \emph{modality gap} in CLIP-like models and such gap has a significant impact on model generalization ability. To this end, inspired by \emph{denoising} auto-encoders \cite{vincent2008extracting}, we propose to train MultiCapCLIP with augmented text features $f'_t$. Here we consider both the input augmentation (IA) and the feature augmentation (FA). Specifically, IA replaces the source text $S$ with a semantically similar one $S'$ to obtain $f'_t$: 
\begin{equation}
\label{eq:global_feature_IP}
   f'_t = {\rm Norm}(\phi_t(S')),
\end{equation}
where $S' \sim \mathbb{X}_S$ and $\mathbb{X}_S = \{S, S'_1, \dots, S'_{N-1}\}$ denotes the candidate set of $S$. For simplicity, we use $\phi_t(\cdot)$ to measure similarities among text in the dataset and select the most similar $N-1$ text to construct $\mathbb{X}_S$ for each $S$. Since we sample text from $\mathbb{X}_S$ with uniform probability, there will be $1/N$ probability that the input text keeps unchanged. As for FA, we follow \citet{li2021simple} to add Gaussian noise $n \sim \mathcal{N}(0, \epsilon)$ into text features. Hence, Eq.~(\ref{eq:global_feature_IP}) can be further extended to:
\begin{equation}
\label{eq:global_feature_IP_FP}
   f'_t = {\rm Norm}({\rm Norm}(\phi_t(S')) + n).
\end{equation}
During training, we replace $f_t$ in Eq.~(\ref{eq:text_feature}) with $f'_t$ in Eq.~(\ref{eq:global_feature_IP_FP}) to encourage the model to learn more robust latent representations.

\section{Main Experiments}

In this section, we first introduce the datasets, metrics, settings of the experiments; Then, we provide the out-of-domain and in-domain results of our approach for zero-shot visual captioning.

\subsection{Experimental Setups}
\label{sec:experimental_setup}

\begin{table}[t]
    \centering
    \footnotesize
    \setlength\tabcolsep{4pt}
    \begin{tabular}{lccccc}  
    \toprule
    &\textbf{MS-COCO} &\textbf{MSR-VTT} &\textbf{VATEX}\cr
    \midrule
    Vision Type         &Image      &Video      &Video\cr
    \midrule
    Training size       &113,287    &6,513      &25,006\cr
    Validation size     &5,000      &497        &1,393\cr
    Testing size        &5,000      &2,990      &1,500\cr
    \midrule
    \# Captions         &616,767    &200,000    &278,990\cr
    Avg. Length         &10.6       &9.3        &12.3\cr
    Target Language     &English    &English    &Chinese\cr
    \bottomrule
    \end{tabular}
    \caption{Statistics of datasets used in main experiments.}
    \label{tab:statistics} 
\end{table}

\paragraph{Datasets.} As shown in Table~\ref{tab:statistics}, we use three benchmark datasets under CC BY 4.0 licence in this section: MS-COCO \cite{chen2015microsoft}, MSR-VTT \cite{xu2016msr}, and VATEX \cite{wang2019vatex}. We apply the \citeposs{karpathy2015deep} split to MS-COCO and follow the official split of MSR-VTT for English captioning. Besides, VATEX is a multilingual video captioning dataset that contains parallel English-Chinese captions. We use it for Chinese captioning\footnote{The official splits of VATEX is 25,991: 3,000: 6,000. However, some video clips are no longer available, resulting in the splits 25,006: 2,893: 5,792. Besides, VATEX does not provide Chinese captions of the test set (\url{https://eric-xw.github.io/vatex-website}). We construct new validation and test sets from the original validation set.}. In Section~\ref{sec:analysis}, we will further use the Multi30K dataset \cite{elliott-etal-2016-multi30k} for German and French caption generation.

\paragraph{Metrics.} Following the common practice in the literature, we report BLEU@4 \cite{papineni-etal-2002-bleu}, METEOR \cite{banerjee-lavie-2005-meteor}, ROUGE-L \cite{lin-2004-rouge} and CIDEr \cite{vedantam2015cider} for video captioning, and additionally measure SPICE \cite{anderson2016spice} for image captioning. All metrics are computed by Microsoft COCO Evaluation Server\footnote{For non-English captioning, we do not report METEOR and SPICE metrics because their implementations consider synonym matching and named entity recognition in English.} \cite{chen2015microsoft}.

\begin{table}[t]
\centering
\footnotesize
\setlength\tabcolsep{3pt}
    
\begin{tabular}{@{}clcc@{}}
\toprule
 
 \multicolumn{2}{c}{\begin{tabular}[c]{@{}c@{}} \bf Settings and \\ \bf Languages \end{tabular}} & \begin{tabular}[c]{@{}c@{}} \bf Training Data + Prompts
 \\ \bf(\textit{Text-only data}) \end{tabular}
&  \bf  Testing Data   \\
\midrule

\multirow{4}{*}{\rotatebox{90}{\begin{tabular}[c]{@{}c@{}}Out-of-\\Domain \end{tabular}}} & \multirow{2}{*}[0pt]{English} & MSR-VTT &  MS-COCO \\
& & MS-COCO & MSR-VTT \\
\cmidrule(){2-4}
& Chinese & MSR-VTT-CN & VATEX \\

\cmidrule(){1-4}
\multirow{4}{*}{\rotatebox{90}{\begin{tabular}[c]{@{}c@{}}In-\\Domain \end{tabular}}} & \multirow{2}{*}[0pt]{English} & MS-COCO &  MS-COCO \\
& & MSR-VTT & MSR-VTT \\
\cmidrule(){2-4}
& Chinese & VATEX & VATEX \\
\bottomrule
\end{tabular}
    \caption{Training and testing data used for different experimental settings. We adopt two English captioning datasets: MS-COCO \cite{chen2015microsoft}, MSR-VTT \cite{xu2016msr} and two Chinese captioning datasets: MSR-VTT-CN \cite{wang2022cross}, and VATEX \cite{wang2019vatex}, to conduct the main experiments.}
    \label{tab:settings}
\end{table}


\begin{table*}[t]
    \centering  
    \scriptsize
    \setlength{\tabcolsep}{2.5pt}
    \begin{tabular}{c l c cc ccccc cccc ccc}
    \toprule
    
    \multirow{2}{*}[-3pt]{\bf Settings}  & \multirow{2}{*}[-3pt]{\bf Methods}  
    &\multirow{2}{*}[-3pt]{\bf Pre-trained Backbone} 
    &\multicolumn{2}{c}{\bf Training Data}
    &\multicolumn{5}{c}{\bf MS-COCO (English)}
    &\multicolumn{4}{c}{\bf MSR-VTT (English)}
    &\multicolumn{3}{c}{\bf VATEX (Chinese)}
    \\
    \cmidrule(lr){4-5}
    \cmidrule(lr){6-10}
    \cmidrule(lr){11-14}
    \cmidrule(lr){15-17}
    &&&Vision &Text &B@4&M&R-L&C&S
    &B@4&M&R-L&C
    &B@4&R-L&C\\
    \midrule

    \multirow{5}{*}{\rotatebox{90}{\begin{tabular}[c]{@{}c@{}}  Weakly-\\Supervised\end{tabular}}} & UIC$\dagger$ \citeyearpar{feng2019unsupervised}
    &Inception + Faster R-CNN &$\surd$ &$\surd$
    &5.6&12.4&28.7&28.6&8.1&-&-&-&-&-&-&-\\
    
    & IC-SME$\dagger$ \citeyearpar{laina2019towards}
    &ResNet + Faster R-CNN &$\surd$ &$\surd$
    &6.5&12.9&35.1&22.7&-&-&-&-&-&-&-&-\\

    & R$^2$M$\dagger$ \citeyearpar{guo2020recurrent}
    &Faster R-CNN &$\surd$ &$\surd$
    &6.4 &13.0 &31.3 &29.0 &\textbf{9.1}&-&-&-&-&-&-&-\\

    & TSGAN$\dagger$ \citeyearpar{zhou2021triple}
    &Faster R-CNN &$\surd$ &$\surd$
    &6.9 &13.0 &32.3 &28.9 &8.3&-&-&-&-&-&-&-\\
    
    & SGM$\dagger$ \citeyearpar{honda-etal-2021-removing}
    &Inception + Faster R-CNN &$\surd$ &$\surd$
    &6.3&14.0&34.5&\textbf{31.9}&8.6&-&-&-&-&-&-&-\\
    \midrule

    \multirow{5}{*}{\rotatebox{90}{Zero-Shot}}  & ZeroCap \citeyearpar{tewel2022zerocap}
    &CLIP + GPT &\xmark &\xmark
    &2.6    &11.5   &-      &14.6   &5.5
    &2.3    &12.9   &30.4   &5.8    &-&-&-\\

    & MAGIC
    \citeyearpar{su2022language}
    &CLIP + GPT &\xmark &$\surd$
    &5.2   &12.5   &30.7   &18.3   &5.7
    &5.5    &13.3   &35.4   &7.4    &-&-&-\\

    & EPT \citeyearpar{tewel2022zero}
    &CLIP + GPT &\xmark &\xmark
    &-&-&-&-&-
    &3.0    &14.6   &27.7   &11.3
    &-&-&-\\

    & ZS-CapCLIP\ssymbol{1} \citeyearpar{radford2021learning}
    & CLIP &\xmark &$\surd$
    &3.4    &13.0   &27.6   &12.2   &6.2
    &4.0    &15.0   &31.0   &5.0
    &2.8    &25.8   &2.0\\

    &  \cellcolor{gray!10}MultiCapCLIP (\textbf{Ours})
    &\cellcolor{gray!10}CLIP &\cellcolor{gray!10}\xmark &\cellcolor{gray!10}$\surd$
    &\cellcolor{gray!10}\textbf{9.7}  &\cellcolor{gray!10}\textbf{15.8}  &\cellcolor{gray!10}\textbf{37.6}  &\cellcolor{gray!10}30.2  &\cellcolor{gray!10}8.9
    &\cellcolor{gray!10}\textbf{13.3}  &\cellcolor{gray!10}\textbf{19.5}  &\cellcolor{gray!10}\textbf{43.3}  &\cellcolor{gray!10}\textbf{15.5}
    &\cellcolor{gray!10}\bf 8.4   &\cellcolor{gray!10}\bf 31.2   &\cellcolor{gray!10}\bf 6.2\\

    \bottomrule
    \end{tabular}
    \caption{\textbf{Out-of-domain visual captioning results}. B@4, M, R-L, C, and S are short for BLEU@4, METEOR, ROUGE-L, CIDEr, and SPICE, respectively. $\dagger$: Training with a corpus with more than 2.3M sentences.  \ssymbol{1} denotes our implementations. All previous works can not deal with multilingual zero-shot captioning. Our approach achieves the best results on most metrics across three datasets.
    }
    \label{tb:main_result_out_of_domain}
\end{table*}

\paragraph{Settings}
As shown in Table~\ref{tab:settings}, we conduct the out-of-domain and in-domain experiments.
1) \textit{Out-of-Domain Experiments} are performed by training the model on the text-only data of A dataset, and then evaluating on the B dataset.
2) \textit{In-Domain Experiments} are conducted by training the model on the text-only data of A dataset, and then evaluating on the A dataset.

\paragraph{Baselines}
Since previous works can not generate zero-shot multilingual visual captions directly, we implement a zero-shot CLIP-based model: \textit{ZS-CapCLIP}, which is trained on text-only data with the same architecture as our MultiCapCLIP but without our proposed concept prompts and text augmentations. To observe the gap between zero-shot and fully-supervised methods, We also implement \textit{CapCLIP} trained on vision-caption pairs.

\paragraph{Implementations.} 
Following the previous works in zero-shot captioning \cite{tewel2022zerocap,su2022language,zeng2023socratic}, we adopt the CLIP (ViT-B/16 variant) \cite{radford2021learning} as our image encoder and text encoder, and adopt a randomly initialized Transformer-BASE \cite{Vaswani2017transformer} as our language decoder. We adopt the same vocabulary as BERT / multilingual BERT \cite{devlin-etal-2019-bert} for English / non-English captioning.
We use the Jieba toolkit\footnote{\url{https://github.com/fxsjy/jieba}} to segment Chinese sentences. We select the hyperparameter $K$ from values $\{4, 8, 16, 32\}$, $N$ from values $\{5, 10, 20\}$ and $\epsilon$ from values $\{0.01, 0.1, 1.0\}$ according to the CIDEr performance on the validation split, and set $K=16$, $N=5$, $\epsilon = 0.01$ for all datasets and settings except that $\epsilon = 0.1$ for in-domain experiments on MS-COCO. 
During training, we apply label smoothing \cite{szegedy2016rethinking} of 0.1, use batches of 32 samples and AdamW \cite{loshchilov2019decoupled} with L2 weight decay of 0.01 to train models for 10 epochs. We set the learning rate fixed to 1e-4 with 10\% warm-up iterations when training on text-only data.
During inference, we use beam search with a beam size of 3 to generate captions.

\subsection{Out-of-Domain Results}
In this section, we evaluate the zero-shot  multilingual captioning performance of our approach under out-of-domain settings. We can notice from Table~\ref{tb:main_result_out_of_domain} that our zero-shot model MultiCapCLIP achieves competitive performance on three datasets across English and Chinese. Although SGM \cite{honda-etal-2021-removing} and R$^2$M \cite{guo2020recurrent} perform better than our model on CIDEr and SPICE metrics on MS-COCO, they require the large-scale image datasets for training and use a larger training corpus (2.3M sentences) than ours (130K sentences). While the previous methods do not target non-English caption generation, our MultiCapCLIP gains obvious relative improvements against the CapCLIP on VATEX Chinese captioning.
The out-of-domain results show that our approach is able to generate multilingual visual captions without any labeled vision-caption data, which could have the potential to promote the application of visual captioning for low-resource language applications.

\begin{table*}[t]
    \centering  
    \fontsize{8.5}{9}\selectfont
    \setlength{\tabcolsep}{4pt}
    \begin{tabular}{c l ccccc cccc ccc}
    \toprule
    
    \multirow{2}{*}[-3pt]{\bf Settings}
    &\multirow{2}{*}[-3pt]{\bf Methods}
    &\multicolumn{5}{c}{\bf MS-COCO (English)}
    &\multicolumn{4}{c}{\bf MSR-VTT (English)} &\multicolumn{3}{c}{\bf VATEX (Chinese)}
    \\
    \cmidrule(lr){3-7}
    \cmidrule(lr){8-11}
    \cmidrule(lr){12-14}
    &&B@4&M&R-L&C&S
    &B@4&M&R-L&C
    &B@4&R-L&C\\
    \midrule
    \multirow{11}{*}{\rotatebox{90}{\begin{tabular}[c]{@{}c@{}}  Fully-\\Supervised \end{tabular}}}
    
    &VATEX \citeyearpar{wang2019vatex}
    &-&-&-&-&-
    &-&-&-&-
    &23.4  &46.0   &39.4\\
    
    &MAD+SAP \citeyearpar{huang2020image} 
    &37.0   &28.1   &57.2   &117.3  &21.3
    &41.3   &28.3   &61.4   &48.5
    &-      &-      &-\\

    &Oscar$_{\rm base}$ \citeyearpar{li2020oscar}
    &36.5   &30.3   &-      &123.7  &23.1
    &-      &-      &-      &-
    &-      &-      &-\\

    &OpenBook \citeyearpar{zhang2021open}
    &-      &-      &-      &-      &-
    &42.8   &29.3   &61.7   &52.9
    &-      &-      &-\\

    &ClipCap \citeyearpar{mokady2021clipcap}
    &33.5   &27.5   &-      &113.1  &21.1
    &-      &-      &-      &-
    &28.3\ssymbol{1}   &49.5\ssymbol{1}   &51.3\ssymbol{1}\\

    & CapCLIP\ssymbol{1} \citeyearpar{radford2021learning}
    &32.3   &27.7   &55.4   &109.5  &20.7
    &42.9   &29.8   &62.3   &54.5
    &29.7   &49.8   &51.0\\

    &CaMEL \citeyearpar{barraco2022camel}
    &39.1   &29.4   &58.5   &125.7  &22.2
    &-      &-      &-      &-
    &-      &-      &-\\
    
    &LEMON$_{\rm base}$ \citeyearpar{hu2022scaling} 
    &40.3   &30.2   &-      &133.3  &23.3
    &-      &-      &-      &-
    &-      &-      &-\\
    
    &SwinBERT \citeyearpar{lin2022swinbert}
    &-      &-      &-      &-      &-
    &45.4   &30.6   &64.1   &55.9
    &-      &-      &-\\
    
    &CLIP-DCD \citeyearpar{yang2022clip} &-&-&-&-&-
    &48.2   &31.3   &64.8   &58.7
    &-      &-      &-\\

    &MV-GPT \citeyearpar{seo2022end}
    &-&-&-&-&-
    &48.9   &38.7   &64.0   &60.0
    &-      &-      &-\\

    &GIT \citeyearpar{wang2022git}    
    &44.1   &31.5   &-      &144.8  &24.7
    &53.8   &32.9   &67.7   &73.9
    &-      &-      &-\\

    \midrule

    \multirow{7}{*}{\rotatebox{90}{\begin{tabular}[c]{@{}c@{}}  Weakly-\\Supervised \end{tabular}}} 
    &UIC \citeyearpar{feng2019unsupervised}&18.6&17.9&43.1&54.9&11.1&-&-&-&-&-&-&-\\
    &IC-SME \citeyearpar{laina2019towards}&19.3&20.2&45.0&61.8&12.9&-&-&-&-&-&-&-\\
    &Graph-Align \citeyearpar{gu2019unpaired} &21.5 &20.9 &47.2 &69.5 &15.0&-&-&-&-&-&-&-\\
    &IGGAN \citeyearpar{cao2020interactions} &21.9 &21.1 &46.5 &64.0 &14.5&-&-&-&-&-&-&-\\
    &TSGAN \citeyearpar{zhou2021triple} &18.9 &18.2 &43.3 &55.2 &11.3&-&-&-&-&-&-&-\\
    &USGAE \citeyearpar{yang2022auto} &17.1 &19.1 &43.8 &55.1 &12.8&-&-&-&-&-&-&-\\
    &SCS \citeyearpar{ben2022unpaired} &22.8 &21.4 &47.7 &74.7 &15.1&-&-&-&-&-&-&-\\
    
    \midrule
    
    \multirow{3}{*}{\rotatebox{90}{\begin{tabular}[c]{@{}c@{}}  Zero-\\Shot \end{tabular}}}  & MAGIC \citeyearpar{su2022language}
    &12.9   &17.4   &39.9   &49.3   &11.3
    &-&-&-&-
    &-&-&-\\

    & ZS-CapCLIP\ssymbol{1} \citeyearpar{radford2021learning}
    &6.1    &15.8   &33.0   &27.3   &9.3
    &8.6    &19.8   &37.3   &11.1
    &21.2   &45.0   &31.8 \\
    
    &\cellcolor{gray!10}MultiCapCLIP (\textbf{Ours})
    &\cellcolor{gray!10}\textbf{27.6}  &\cellcolor{gray!10}\textbf{25.2}  &\cellcolor{gray!10}\textbf{51.6}  &\cellcolor{gray!10}\textbf{96.2}  &\cellcolor{gray!10}\textbf{18.5}
    &\cellcolor{gray!10}\textbf{22.0}  &\cellcolor{gray!10}\textbf{24.4}  &\cellcolor{gray!10}\textbf{50.2}  &\cellcolor{gray!10}\textbf{33.6}
    &\cellcolor{gray!10}\textbf{22.8}  &\cellcolor{gray!10}\textbf{46.0}  &\cellcolor{gray!10}\textbf{38.2}
    \\

    \bottomrule
    \end{tabular}
    \caption{\textbf{In-domain visual captioning results}. Fully-supervised and large-scale pre-trained models are included for comparisons. \ssymbol{1} denotes our implementations. Our MultiCapCLIP significantly outperforms previous zero-shot/weakly-supervised models, but still suffers from performance gaps compared with fully-supervised counterparts.}
    \label{tab:in-domain}
\end{table*}

\begin{figure*}[t]
\centering
\includegraphics[width=0.94\linewidth]{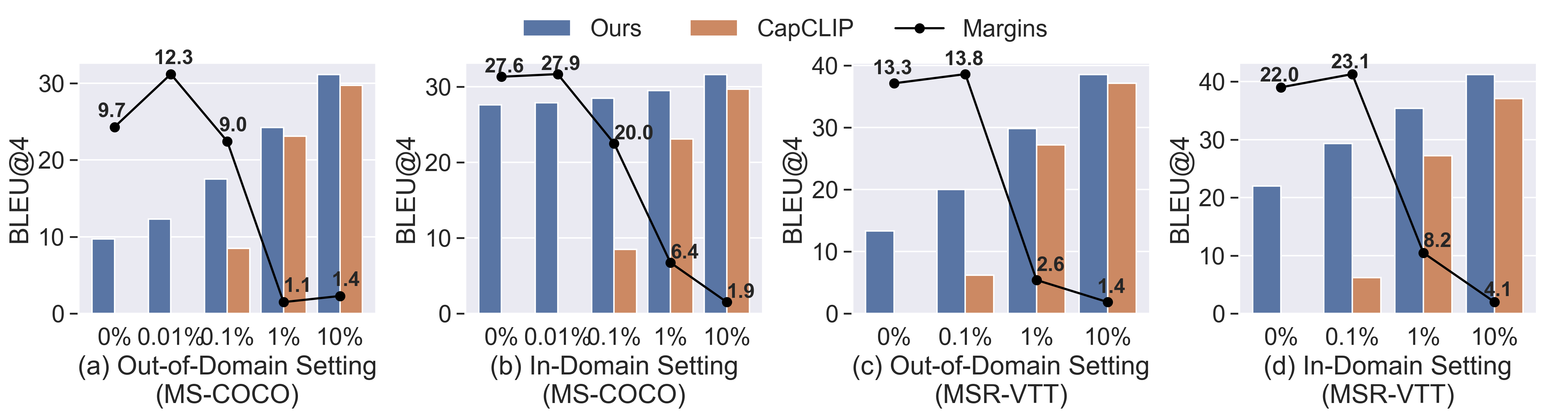}
\caption{Results of out-of-domain and in-domain experiments with respect to different ratios of training data. The margins in different ratios are shown with the polyline. Our method consistently surpasses CapCLIP, and the fewer the vision-caption pairs, the larger the margins.}
\label{fig:semi}
\end{figure*}

\begin{table*}[t]
    \centering   
    \small
    \setlength{\tabcolsep}{4.5pt}
    \begin{tabular}{lccccccccccccccc}
    \toprule
    \multirow{2}{*}[-3pt]{\bf Setting} & \multicolumn{3}{c}{\bf Component} 
    &\multirow{2}{*}[-3pt]{$\bm{K}$}
    &\multirow{2}{*}[-3pt]{\begin{tabular}[c]{@{}c@{}} \bf Concept Type \end{tabular}}
    &\multicolumn{5}{c}{\bf Out-of-Domain Setting}
    &\multicolumn{5}{c}{\bf In-Domain Setting} \\
    \cmidrule(lr){2-4}
    \cmidrule(lr){7-11}
    \cmidrule(lr){12-16}
    & CP &IA &FA
    &&&B@4&M&R-L&C&S
    &B@4&M&R-L&C&S
    \\
    \midrule
    
    Base Model &
    & & & - &-
    &3.4    &13.0   &27.6   &12.2   &6.2
    &6.1    &15.8   &33.0   &27.3   &9.3
    \\
    \midrule
    
    (a)& $\surd$&& &16 &Noun
    &5.7    &15.2   &32.4   &18.1   &8.6
    &7.7    &17.6   &35.9   &40.6   &12.3\\
    
    (b) & &$\surd$& & - &-
    &7.4    &15.3   &34.1   &23.8   &8.5
    &15.1   &20.5   &42.2   &57.6   &14.5\\
    (c) &&&$\surd$ & - &-
    &5.5    &14.3   &32.1   &15.3   &7.3
    &26.1   &25.2   &51.2   &91.5   &18.3\\

    (d) &&$\surd$&$\surd$ &- &-
    &7.4    &15.7   &34.4   &23.9   &\textbf{9.2}
    &26.6   &25.2   &51.3   &92.5   &18.4\\
    \cmidrule(lr){2-16}
    (e) &$\surd$&$\surd$&$\surd$ &4 &Noun 
    &8.2    &15.1   &35.9   &27.7   &8.1
    &27.7	&25.2	&51.9	&94.6	&18.4\\
    (f) &$\surd$&$\surd$&$\surd$ &8 &Noun 
    &8.1    &15.6   &35.3   &29.3   &8.5
    &27.5   &25.1 &51.6 &95.0 &18.3
    \\
    \rowcolor{gray!10}(g) Full Model & $\surd$&$\surd$&$\surd$ &16 &Noun
    &\textbf{9.7}  &15.8  &\textbf{37.6}  &\textbf{30.2}  &8.9
    &27.6  &\textbf{25.2}  &51.6  &\textbf{96.2}  &\textbf{18.5} \\
    (h) &$\surd$&$\surd$&$\surd$ &32 &Noun
    &9.1   &\textbf{16.2}   &37.1   &30.1   &9.1
    &\textbf{28.4}	&\textbf{25.2}	&\textbf{51.9}	&95.7	&\textbf{18.5}\\

    \cmidrule(lr){2-16}
    (i) & $\surd$&$\surd$&$\surd$ &16 &Verb
    &7.0 &15.0 &34.1 &21.1 &7.1
    &27.8 &\textbf{25.2} &\textbf{51.9} &93.2 &18.3\\
    (j) & $\surd$&$\surd$&$\surd$ &16 &Noun + Verb 
    &9.2  &15.7 &37.0 &28.4 &8.6
    &27.1 &25.1 &51.4 &94.3 &18.5\\
    \bottomrule
    
    \end{tabular}
    \caption{Quantitative analysis of the proposed MultiCapCLIP, which includes visual concept prompts (CP), input augmentation (IA), and feature augmentation (FA). We conduct the ablation study on the MS-COCO dataset under out-of-domain and in-domain settings. The full model denotes our proposed MultiCapCLIP.}
    \label{tab:ablation_main}
\end{table*}

\subsection{In-Domain Results}
For comparisons, we further consider state-of-the-art fully-supervised and large-scale pre-trained models and models under the \emph{unpaired} setting, i.e., both vision and text data of the target dataset are utilized for training independently, leaving their pairing annotations unused. As shown in Table~\ref{tab:in-domain}, our approach significantly outperforms previous unpaired/zero-shot competitors by up to 4.8\% BLEU@4, 3.9\% ROUGE-L, and 21.5\% CIDEr scores in MS-COCO English captioning. When it comes to MSR-VTT English captioning and VATEX Chinese captioning, our MultiCapCLIP surpasses ZS-CapCLIP by a large margin under the CIDEr metric, e.g., an absolute improvement of 22.5\% on MSR-VTT. These results prove the effectiveness of MultiCapCLIP in zero-shot multilingual visual captioning. Nevertheless, there still exists performance gaps between MultiCapCLIP trained on text-only data and existing state-of-the-art fully-supervised models trained on full vision-text data. 

\section{Analysis}
\label{sec:analysis}
In this section, we conduct several analyses to better understand our approach.

\subsection{Semi-Supervised Visual Captioning}
To further prove the effectiveness of our approach, we fine-tune MultiCapCLIP with partial labeled vision-caption data of downstream datasets. To this end, in Figure~\ref{fig:semi}, we evaluate the performance of MultiCapCLIP with respect to the increasing amount of labeled data. Specifically, we randomly sample a small portion of training images/videos and use the resulting vision-caption pairs for fine-tuning. We repeat this process by three times and report the average performance. For a fair comparison, we also train CapCLIP (Section~\ref{sec:experimental_setup}) with the same amount of pairs. As we can see in Figure~\ref{fig:semi}, for both in-domain or a out-of-domain corpus, MultiCapCLIP consistently outperforms CapCLIP with different ratios of training data. It is worth noting that the fewer the labeled vision-caption pairs, the larger the margins. In detail, under the extremely low label setting, e.g., 0.1\% of paired data on MSR-VTT (only 6 videos), our approach under the in-domain setting significantly surpasses the CapCLIP by 23.1\% absolute BLEU@4 score. It further proves the effectiveness of our approach, which can relax the reliance on the vision-caption annotations. We can make use of available unpaired text-only data as a solid basis for multilingual visual captioning tasks. 

\subsection{Quantitative Analysis}
In this section, we analyze the contributions of each component in our approach.

\paragraph{Ablation Study} 
We conduct the ablation study on the out-of-domain and in-domain settings using MS-COCO dataset \cite{chen2015microsoft}.
As shown in Table~\ref{tab:ablation_main}, each component in our proposed approach can boost the performance over all metrics, verifying our arguments and the effectiveness of our approach. 
In particular, setting (a) shows that the introduced prompts can improve the base model with absolute gains up to 5.9\% and 13.3\% CIDEr scores under out-of-domain and in-domain settings, respectively. 
Settings (b,c) show that either input augmentation (IA) or feature augmentation (FA) respectively boost performance, indicating the importance of bridging the modality gap between the visual and textual data and in turn, boosting the robustness of the model and improving the performance of zero-shot visual captioning. 
Moreover, by comparing the results of (b) and (c), we observe that FA brings more improvements under the in-domain setting whereas IA is better under the out-of-domain setting. 
This indicates that structure noises are more suitable to bridge the modality gap between vision and text data from the same domain. 
From another perspective, we need a more complex feature adaptation method for out-of-domain transfer. 
Since the IA and FA can improve the performance from different perspectives, as shown in setting (d), combining them can lead to the most prominent improvement across all metrics. 
Moreover, compared with (d), our full model in the setting (g) can still gain improvements under most metrics, especially the CIDEr metric, showing that concept prompts benefit visual captioning by generating more accurate details. 

\paragraph{Effect of $K$} 
As shown in Table~\ref{tab:ablation_main} (e-h),  when we set the number of prompts $K$ to 16, the model substantially performs the best. For other $K$ values, when $K < 16$, the performance is improved with an increasing $K$ due to more adequate guidance signals to the model. However, when $K > 16$, we can observe saturated or impaired captioning performance, possibly because retrieving more prompts do not include additional useful clues and introduce irrelevant noises to the model.

\paragraph{Concept Type} Other than prompting the model with noun phrases (Section \ref{sec:concept_prompts}), we also consider the effect of verbs. As shown in Table~\ref{tab:ablation_main}, setting (g) surpasses settings (i) and (j) at most cases, i.e., using verb-based prompts degrades performance. We speculate the reason is that the vision-language model we used (i.e., CLIP) can recognize salient objects more accurately than human actions. 

\begin{figure*}[t]
\centering
\includegraphics[width=1\linewidth]{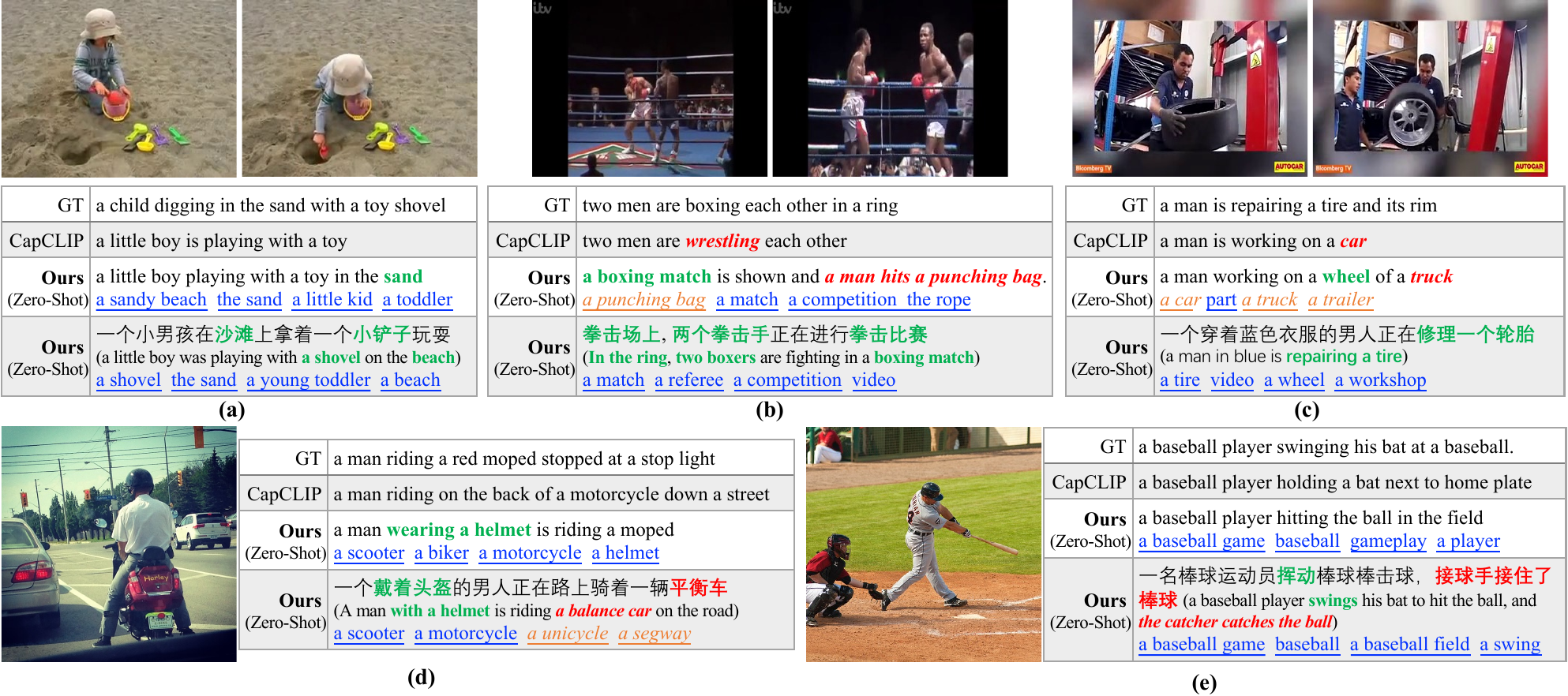}
\caption{Captioning comparisons between ground-truths (GT), CapCLIP trained with full in-domain vision-text pairs, and our zero-shot MultiCapCLIP trained with out-of-domain corpora. We emphasize \textcolor{darkgreen}{\textbf{accurate}} and \textcolor{red}{\emph{\textbf{wrong}}} keywords and highlight \textcolor{blue}{\underline{reasonable}} and \textcolor{orange}{\underline{\emph{noisy}}} concepts used for prompting. Our approach can generate plausible visual descriptions in English and Chinese without the need of vision-caption pairs.}

\label{fig:example}
\end{figure*}

\subsection{Robustness Analysis: Extensions to More Non-English Languages} We adopt the Multi30K dataset \cite{elliott-etal-2016-multi30k} to further evaluate in-domain performance on German and French image captioning. As shown in Table~\ref{tab:ablation_lang}, our full model again outperforms the base model by a large margin, proving the effectiveness of concept prompts and text augmentations.

\begin{table}[t]
    \centering   
    \small
    \setlength{\tabcolsep}{4pt}
    \begin{tabular}{lcccccc}
    \toprule
    \multirow{2}{*}[-3pt]{\bf Setting}
    &\multicolumn{3}{c}{\bf German}
    &\multicolumn{3}{c}{\bf French} \\
    \cmidrule(lr){2-4}
    \cmidrule(lr){5-7}

    &B@4&R-L&C
    &B@4&R-L&C
    \\
    \midrule
    Supervised &20.0 &45.7 &55.8 &7.1 &28.0 &54.0\\
    \midrule
    ZS-Base Model &3.8 &27.7 &10.7 &2.6 &19.4 &20.4\\
    \rowcolor{gray!10}ZS-Full Model &\textbf{13.3} &\textbf{38.3} &\textbf{36.7} &\textbf{5.2} &\textbf{23.9} &\textbf{40.5}\\
    \bottomrule
    
    \end{tabular}
    \caption{In-domain performance on German and French image captioning. ZS is short for ``Zero-Shot''.}
    \label{tab:ablation_lang}
\end{table}

\subsection{Qualitative Analysis}
In this section, we give some visualization results and examples to better understand our approach.

\begin{figure}[t]
\centering
\includegraphics[width=0.99\linewidth]{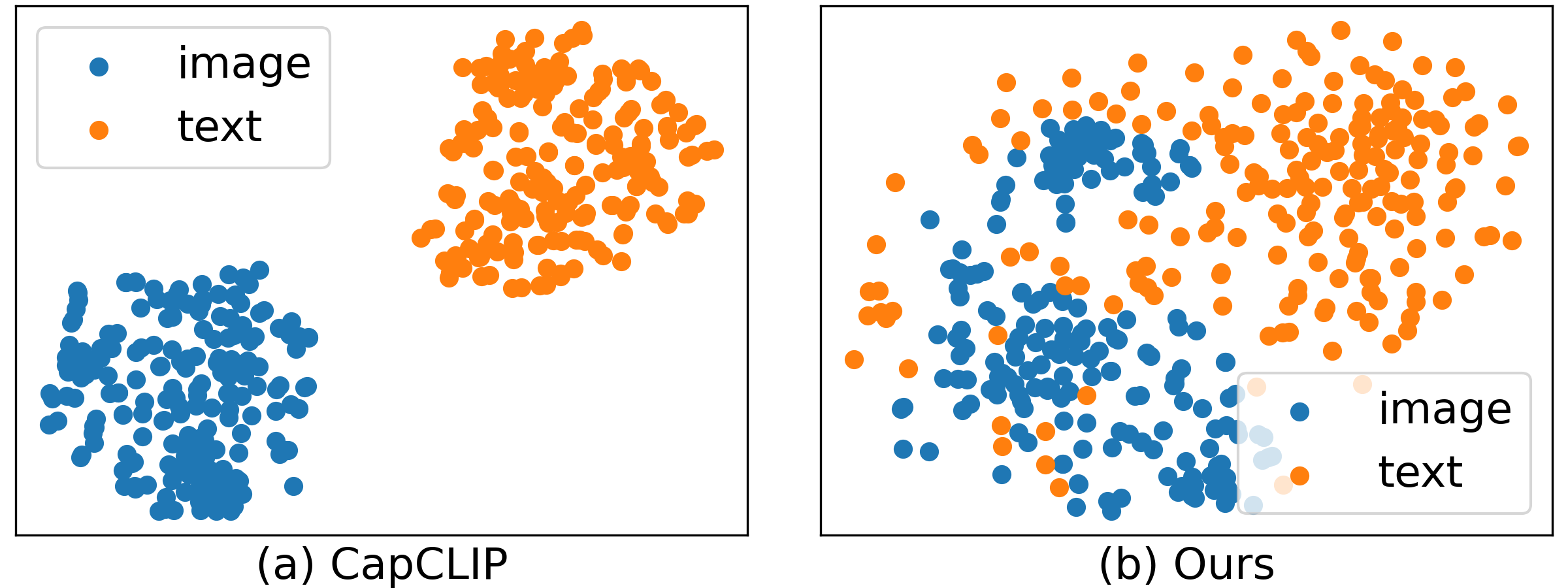}
\caption{T-SNE visualization \cite{tsne} of image and text embeddings produced by (a) ZS-CapCLIP and (b) our MultiCapCLIP during training. We plot the scatter diagrams for 200 image-caption pairs. Our approach can effectively bridge the gap between the vision and text modalities.}
\label{fig:vis}
\end{figure}

\paragraph{Visualization} To verify the effect of our method on representation learning, we use t-SNE \cite{tsne} to visualize the features produced by ZS-CapCLIP and our MultiCapCLIP in Figure~\ref{fig:vis}, which shows that our approach can bridge the modality gap between visual and textual inputs during training and obtain a blended distribution, leading to a more robust shift from text-to-text generation to vision-to-text generation.

\paragraph{Examples}
In Figure~\ref{fig:example}, we compare our model trained with out-of-domain corpora with CapCLIP trained on full in-domain supervision. As we can see, our model can generate accurate keywords, e.g., ``sand'' in (a), ``tire'' in (c), and ``helmet'' in (d), which can be attributed to the useful clues of concept prompts. However, there exist noises in the retrieved concepts in some cases, e.g., ``a punching bag'' in (b), misleading the model to produce wrong details. Besides, in (e), we can observe how the training corpus affect the writing style of the model: the corpus of a video caption dataset (VATEX) makes the model focus on the temporal evolution of events, resulting in a speculative description ``the catcher catches the ball''. Overall, our approach can be a solid basis for zero-shot multilingual visual captioning. It requires no vision-caption pairs but generates plausible visual descriptions.

\section{Related Works}
The related works are introduced from zero-shot learning and visual captioning.

\paragraph{Zero-shot Learning}

Adapting models to novel tasks with limited labeled data is an important research topic toward general intelligence \cite{griffiths2019doing}. 
Contrastive pre-training is an effective technique to achieve this goal and has revolutionized multimodal research \cite{hou2021exploring,gan2022vision,jin2022expectation,cheng2023}. 
Specifically for the vision-language field, models such as CLIP \cite{radford2021learning} and ALIGN \cite{jia2021scaling} learn a shared multimodal embedding space from large-scale noisy image-text pairs, leading to an impressive zero-shot performance on tasks like image classification and vision-text retrieval \cite{2022tip,luo2022clip4clip}. Nevertheless, employing CLIP-like models in low-data vision-grounded text generation (i.e., visual captioning) remains challenging.


\paragraph{Visual Captioning} 
As a key vision-language task, visual captioning has achieved tremendous progress under the encoder-decoder framework \cite{xu2015show} and the ``pre-training and fine-tuning'' paradigm. Yet, typical visual captioning methods require curated datasets of numerous images or videos paired with descriptions in a specific language, which are costly to collect. To this end, some weakly-supervised approaches are proposed \cite{feng2019unsupervised,guo2020recurrent,honda-etal-2021-removing,ben2022unpaired}. These methods require disjoint vision and text data for training and rely on a pre-trained object detector like Faster R-CNN \cite{ren2015faster} to construct weak supervision signals. However, the detectors they use are limited to a pre-defined set of categories. Recently, several works integrate CLIP with large language models (LLMs) like GPT \cite{radford2019language,brown2020language} for zero-shot visual captioning \cite{tewel2022zerocap,su2022language,liu2022aligning,zeng2023socratic}. 
Although effective, these methods suffer from over-parameterization of large LLMs. We instead train a lightweight decoder from scratch. Besides, some concurrent works address zero-shot visual captioning by training CLIP with text-only data \cite{nukrai2022text,gu2022can,li2023decap,yang2023zeronlg}. What differentiates our work from them is that we consider visual concept prompts that perverse domain visual knowledge.

\section{Conclusions}
We have presented a data-efficient method dubbed MultiCapCLIP to re-purpose CLIP-like vision-language pre-trained models for zero-shot multilingual visual captioning. Our approach reduces the reliance on labeled vision-caption pairs of downstream datasets by auto-encoding concept prompts on text-only data. Extensive experiments on four datasets and four languages confirm the effectiveness of our approach, which can be a solid basis for visual captioning in low-data regimes and low-resource languages.

\section*{Limitations}
Although the proposed MultiCapCLIP can generate multilingual zero-shot visual captions without any labeled vision-caption training pairs.
We still need the independent set of text for training/translating, which may still be difficult to collect for some low-resource languages. 
This might be alleviated in the future with techniques such as knowledge distillation from publicly-available pre-trained models, e.g., BERT \cite{devlin-etal-2019-bert}. Besides, our approach uses CLIP to measure text-text similarities for retrieving concept prompts and conducting input augmentation during training. Considering that CLIP is optimized by image-text global contrast \cite{radford2021learning} and intra-modal retrieval of such a model is not as well as its cross-modal retrieval \cite{jia2021scaling}, an improvement direction of our approach is using a vision-language pre-trained model that measures intra-modal and inter-modal semantic similarities well \cite{yang2022vision}.

\section*{Ethics Statement}
We conduct the experiments on public datasets, which are exclusively about natural images, videos, and captions. These datasets have been carefully pre-processed for the academic study purpose, and therefore do not contain any information that names or uniquely identifies individual people or offensive content. It is noteworthy that our approach inherits the drawback of the pre-trained backbone, i.e., CLIP, which has demonstrated that improper class design used for prompting may raise unwanted biases \cite{radford2021learning}. Therefore, careful examination is needed before employing our approach in real-world scenarios to avoid prejudices.

\section*{Acknowledgements}
This paper was partially supported by NSFC (No: 62176008) and Shenzhen Science \& Technology Research Program ({\small No: GXWD20201231165807007-20200814115301001}).


\bibliography{anthology,custom}
\bibliographystyle{acl_natbib}




\end{document}